\pdfoutput=1

\documentclass[11pt]{article}
\usepackage[table]{xcolor}

\usepackage{acl}

\usepackage{times}
\usepackage{latexsym}

\usepackage[T1]{fontenc}

\usepackage[utf8]{inputenc}

\usepackage{microtype}

\usepackage{url}
\usepackage{easyReview}

\usepackage{cleveref}
\usepackage{multirow}
\usepackage{tabularx}

\usepackage{graphicx}

\usepackage{subfig}
\usepackage{booktabs}

\definecolor{cadetgrey}{rgb}{0.57, 0.64, 0.69}
\definecolor{grey}{rgb}{0.169,0.169,0.169}

\title{\textit{Probing with Noise}: \\ Unpicking the Warp and Weft of Embeddings}

\author{Filip Klubička \and John D. Kelleher \\
  ADAPT Centre, Technological University Dublin, Ireland \\
  \texttt{\{filip.klubicka,john.kelleher\}@adaptcentre.ie} \\}

\begin{document}
\maketitle
\begin{abstract}
Improving our understanding of how information is encoded in vector space can yield valuable interpretability insights. Alongside vector dimensions, we argue that it is possible for the vector norm to also carry linguistic information. We develop a method to test this: an extension of the probing framework which allows for relative intrinsic interpretations of probing results. It relies on introducing noise that ablates information encoded in embeddings, grounded in random baselines and confidence intervals. We apply the method to well-established probing tasks and find evidence that confirms the existence of separate information containers in English GloVe and BERT embeddings. Our correlation analysis aligns with the experimental findings that different encoders use the norm to encode different kinds of information: GloVe stores syntactic and sentence length information in the vector norm, while BERT uses it to encode contextual incongruity.
\end{abstract}

\section{Introduction}

Probing in NLP, as defined by \citet{Conneau:2018}, is a classification problem that predicts linguistic properties using dense embeddings as training data. The framework rests on the assumption that the probe's success at a given task indicates that the encoder is storing information on the pertinent linguistic properties. Probing has quickly become an essential tool for encoder interpretability, by providing interesting insights into embeddings.

In essence, embeddings are vectors positioned in a shared multidimensional vector space, and vectors are geometrically defined by two aspects: having both a \textbf{direction} and \textbf{magnitude} \citep[page 36]{hefferon2018linear}. Direction is the position in the space that the vector points towards (expressed by its dimension values), while magnitude is a vector's length, defined as its distance from the origin (expressed by the vector norm) \citep[page 131]{anton2013elementary}. It is understood that information contained in a vector is encoded in the dimension values, which are most often studied in NLP research (see \S\ref{s:rw}).
However, information can be encoded in a representational vector space in more implicit ways, and relations can be inferred from more than just vector dimension values.

We hypothesise that it is possible for the vector magnitude---the norm---to carry information as well. Though it is a distributed property of a vector's dimensions, the norm not only relates the distance of a vector from the origin, but indirectly also its distance from other vectors. Two vectors could be pointing in the exact same direction, but their distance from the origin might differ dramatically.\footnote{Mathematically, two vectors can only be considered equal if both their direction and magnitude are equal \citep[page 137]{anton2013elementary}.} A similar effect has been observed in the literature: for many word embedding algorithms, the norm of the word vector correlates with the word's frequency \citep{schakel2015measuring}. E.g. in fastText embeddings the vectors of stop words (the most frequent words in English) are positioned closer to the origin than content words \citep{balodis2018intent}; and \citet{goldberg2017neural} notes that for many embeddings normalising the vectors removes word frequency information. Additionally, the norm plays an integral part in BERT's attention layer, controlling the levels of contribution from frequent, less informative words by controlling the norms of their vectors \citep{kobayashi-etal-2020-attention}. It stands to reason that the norm could be leveraged by embedding models to encode other linguistic information as well. Hence, we argue that a vector representation has two \textbf{information containers}: vector \textit{dimensions} and the vector \textit{norm} (the titular \textit{warp} and \textit{weft}). In this paper, we test the assumption that these two components can be used to encode different types of information.

To this end, we need a probing method that provides an intrinsic evaluation of any given embedding representation, for which the typical probing pipeline is not suited. We thus extend the existing probing framework by introducing random noise into the embeddings. This enables us to do an intrinsic evaluation of a single encoder by testing whether the noise disrupted the information in the embedding being tested. The right application of noise enables us to determine which embedding component the relevant information is encoded in, by ablating that component's information. In turn, this can inform our understanding of how certain linguistic properties are encoded in vector space. We call the method \textit{probing with noise} and demonstrate its generalisability to both contextual and static encoders by using it to intrinsically evaluate English GloVe and BERT embeddings on a number of established probing tasks.

This paper's main contributions are: (a) a methodological extension of the probing framework: \textit{probing with noise}; (b) an array of experiments demonstrating the method on a range of probing tasks; and (c) an exploration of the importance of the vector norm in encoding linguistic phenomena in different embedding models. 

\section{Method: Probing With Noise}
\label{s:method}

Our method is an extension of the typical probing pipeline (steps 1-6), incorporated as steps 7 and 8:

\begin{table}[h]
\scalebox{0.99}{
\centering
\begin{tabular}{l}

1. Choose a probing task \\ 
2. Choose or design an appropriate dataset \\
3. Choose a word/sentence representation \\
4. Choose a probing classifier (the probe) \\
5. Train the probe on the embeddings as input \\
6. Evaluate the probe's performance on the task \\
\textbf{7. Introduce systematic noise in the embedding } \\
\textbf{8. Repeat training, evaluate and compare} \\

\end{tabular}}
\label{tab:noiseprobing}
\end{table}

Usually, the evaluation score from step 6 is used as a basis to make inferences regarding the presence of the probed information in embeddings.
Different encoders are compared based on their evaluation score and the probe's relative performance can inform which model stores the information more saliently.
Though ours may seem like a minor addition, it changes the approach conceptually. Now, rather than providing the final score, the output of step 6 establishes an intrinsic, \textit{vanilla baseline}. Embeddings with noise injections can then be compared against it in steps 7 and 8, offering a relative intrinsic interpretation of the evaluation. In other words, using relative information between a vector representation and targeted ablations of itself allows for inferences to be made on where information is encoded in embeddings.

The method relies on three supporting pillars: (a) random baselines, which in tandem with the vanilla baseline provide the basis for a relative evaluation; (b) statistical significance derived from confidence intervals, which informs the inferences we make based on the relative evaluation; and (c) targeted noise, which enables us to examine where the information is encoded. We describe them in the following subsections, starting with the noise.

\subsection{Choosing the Noise}

The nature of the noise is crucial for our method, as the goal is to systematically disrupt the content of the information containers in order to identify whether a container encodes the information. We use an ablation method to do this: by introducing noise into either container we “sabotage” the representation, in turn identifying whether the information we are probing for has been removed. Though we introduce random noise, our choice of how to apply it is systematic, as it is important that the noising function applied to one container leaves the information in the remaining container intact, otherwise the results will not offer relevant insight.

\textbf{Ablating the Dimension Container:} The noise function for ablating the dimensions needs to remove its information completely, while leaving the norm intact. It should also not change the dimensionality of the vector, given that a change in the dimensionality of a feature also changes the chance of the probe finding a random or spurious hyper-plane that performs well on the data sample. Maintaining the dimensionality thus ensures that the probability of the model finding such a lucky split in the feature space remains unchanged. 

Our noise function satisfies these constraints: for each embedding in a dataset, we generate a new, random vector of the same dimensionality, then scale the new dimension values to match the norm of the original vector. This invalidates any semantics assigned to a particular dimension as the values are replaced with meaningless noise, while retaining the original vector's norm values. 

\textbf{Ablating the Norm Container:} To remove information potentially carried by a vector's norm while retaining dimension information, we apply a noising function analogous to the previous one: for each embedding we generate a random norm value, and then scale the vector's original dimension values to match the new norm. This randomises vector magnitudes, while the relative sizes of the dimensions remain unchanged. In other words, all vectors will keep pointing in the same directions, but any information encoded by differences in magnitude is removed.\footnote{We are conscious that vectors have more than one kind of norm, so choosing which norm to scale to might not be trivial. We have explored this in supplementary experiments and found that in our framework there is no significant difference between scaling to the L1 norm vs. L2 norm.}

\textbf{Ablating Both Containers:} The two approaches are not mutually exclusive: applying both noising functions should have a compounding effect and ablate both information containers simultaneously, essentially generating a completely random vector with none of the original information. 

\subsection{Random Baselines}

Even when no information is encoded in an embedding, the train set may contain class imbalance, and the probe can learn the distribution of classes. To account for this, as well as the possibility of a powerful probe detecting an empty signal \citep{zhang-bowman-2018-language}, we need to establish informative random baselines against which we can compare the probe's performance.

We employ two such baselines: (a) we assert a random prediction onto the test set, negating any information that a classifier could have learned, class distributions included; and (b) we train the probe on randomly generated vectors, establishing a baseline with access only to class distributions. 

\subsection{Confidence Intervals}

Finally, we must account for the degrees of randomness, which stem from two sources: (1) the probe may contain a stochastic component, e.g. a random weight initialisation; (2) the noise functions are highly stochastic (i.e. sampling random norm/dimension values). Hence, evaluation scores will differ each time the probe is trained, making relative comparisons of scores problematic. To mitigate this, we retrain and evaluate each model 50 times reporting the average score of all runs, essentially bootstrapping over the random seeds. 

To obtain statistical significance for the averages, we calculate a 99\% confidence interval (CI) to confirm that observed differences in the averages of different model scores are significant. We use the CI range when comparing evaluation scores of probes on any two noise models to determine whether they come from the same distribution: if there is overlap in the range of two possible averages they might belong to the same distribution and there is no statistically significant difference between them. Using CIs in this way gives us a clearly defined decision criterion on whether any model performances are different.

\section{Data}
\label{s:data}

In our experiments we use 10 established probing task datasets for the English language introduced by \citet{Conneau:2018}. The goal of the multi-class \textit{Sentence Length} (SL) probing task is to predict the length of the sentence as binned in 6 possible categories, while \textit{Word Content} (WC) is a task with 1000 words as targets, predicting which of the target words appears in a given sentence. The \textit{Subject} and \textit{Object Number} tasks (SN and ON) are binary classification tasks that predict the grammatical number of the subject/object of the main clause as being singular or plural, while the \textit{Tense} (TE) task predicts whether the main verb of the sentence is in the present or past tense. The \textit{Coordination Inversion} (CIN) task distinguishes between a sentence where the order of two coordinated clausal conjoints has been inverted or not. \textit{Parse Tree Depth} (TD) is a multi-class prediction task where the goal is to predict the maximum depth of the sentence's syntactic tree, while \textit{Top Constituents} (TC) predicts one of 20-classes of the most common syntactic top-constituent sequences.
In the \textit{Bigram Shift} (BS) task, the goal is to predict whether two consecutive tokens in the sentence have been inverted, and \textit{Semantic Odd Man Out} (SOMO) is a task predicting whether a noun or verb was replaced with a different noun or verb.
We use these datasets as published in their totality, with no modifications.\footnote{\url{https://github.com/facebookresearch/SentEval/tree/master/data/probing}} We also consider these tasks to represent examples of different language domains: surface information (SL,WC), morphology (SN,ON,TE), syntax (TD,TC,CIN) and contextual incongruity (BS,SOMO). This level of abstraction can lend itself to interpreting the experimental results, as there may be similarities across tasks in the same domain (note that \citet{durrani-etal-2020-analyzing} follow a similar line of reasoning).

\section{Experiments}
\label{s:experiments}

\subsection{Models and Implementation}

Given the current prominence of contextual encoders, such as BERT \citep{devlin-etal-2019-bert}, ELMo \citep{ELMo} and their derivatives, they are an obvious choice for the application of our method. However, rather than compare different contextual encoders, we prefer to draw a contrastive comparison with a static encoder, such as GloVe \cite{Pennington:2014}, which is a distributed representation based on a word to word co-occurrence matrix. This provides insight into both models and demonstrates the method's generalisability to more than one type of encoder. In our experiments we examine BERT and GloVe embeddings.

Note that all the probing datasets we use are framed as classification tasks at the sentence level (see \S\ref{s:data}), so our experiments require sentence representations. We use pretrained versions of BERT and GloVe to generate embeddings for each sentence. The BERT model generates 12 layers of embedding vectors with each layer containing a separate 768-dimensional embedding for each word, so we average the word embeddings in BERT's final layer, resulting in a 768-dimensional sentence embedding. We take the same mean pooling approach with GloVe, which yields a 300-dimensional sentence embedding for each sentence. While BERT uses sub-word tokens to get around out of vocabulary tokens, in the rare instance of encountering an OOV with GloVe, we generate a random word embedding in its stead.

In each set of experiments, the sentence embeddings are used as input to a Multi-Layered Perceptron (MLP) classifier, which labels them according to the probing task. We evaluate the performance of all probes using the AUC-ROC score.\footnote{\url{https://scikit-learn.org/stable/modules/generated/sklearn.metrics.roc_auc_score.html}} Regarding implementation and parameter details, we used the bert-base-uncased BERT model from the \textit{pytorch\_pretrained\_bert} library\footnote{\url{https://pypi.org/project/pytorch-pretrained-bert/}} \citep{PyTorch}, a pre-trained GloVe model\footnote{The larger common crawl vectors: \url{https://nlp.stanford.edu/projects/glove/}} and for the MLP probe we used the scikit-learn MLP implementation \citep{scikit-learn} using the default parameters.\footnote{activation='relu', solver='adam', max\_iter=200, hidden\_layer\_sizes=100, learning\_rate\_init=0.001, batch\_size=min(200,n\_samples), early\_stopping=False, weight init. $W \sim \mathcal{N}\left(0, \sqrt{6/(fan_{in}+fan_{out})}\right)$ (scikit relu default). See:  \url{https://scikit-learn.org/stable/modules/generated/sklearn.neural_network.MLPClassifier.html} }\textsuperscript{,}\footnote{Code available here: \url{https://github.com/GreenParachute/probing-with-noise}}

\begin{table*}
\centering
\scalebox{0.79}{
\centering
\begin{tabular}{|l|c|c|c|c|c|c|c|c|c|c|c|}
\hline
\multicolumn{11}{|c|}{\textbf{GloVe}} & \textbf{Key} \\
\hline
Model & \multicolumn{2}{m{3.6em}|}{\textbf{SL}} & \multicolumn{2}{m{4.1em}|}{\textbf{WC}} & \multicolumn{2}{m{3.8em}|}{\textbf{SN}} & \multicolumn{2}{m{3.8em}|}{\textbf{ON}} & \multicolumn{2}{m{3.8em}|}{\textbf{TE}} & \textit{Surface Info.} \\
 & auc & \textpm CI & auc & \textpm CI & auc & \textpm CI & auc & \textpm CI & auc & \textpm CI & SL: Sentence Length \\
\cline{1-11}
rand. pred. & \cellcolor{cadetgrey!25} .5006 & \cellcolor{cadetgrey!25} .0013 & \cellcolor{cadetgrey!25} .4995 & \cellcolor{cadetgrey!25} .001 & \cellcolor{cadetgrey!25} .4996 & \cellcolor{cadetgrey!25} .002 & \cellcolor{cadetgrey!25} .4999 & \cellcolor{cadetgrey!25} .0023 & \cellcolor{cadetgrey!25} .4981 & \cellcolor{cadetgrey!25} .0022 & WC: Word Content \\
 rand. vec. & \cellcolor{cadetgrey!25} .4999 & \cellcolor{cadetgrey!25} .0011 & \cellcolor{cadetgrey!25} .5006 & \cellcolor{cadetgrey!25} .0009 & \cellcolor{cadetgrey!25} .499 & \cellcolor{cadetgrey!25} .0022 & \cellcolor{cadetgrey!25} .4998 & \cellcolor{cadetgrey!25} .0024 & \cellcolor{cadetgrey!25} .4997 & \cellcolor{cadetgrey!25} .0024 & \textit{Morphology} \\
\cline{1-11}
 vanilla & \cellcolor{grey!25} .9475 & \cellcolor{grey!25} .0005 & \cellcolor{grey!25} .9974 & \cellcolor{grey!25} .0001 & \cellcolor{grey!25} .8114 & \cellcolor{grey!25} .0014 & \cellcolor{grey!25} .7805 & \cellcolor{grey!25} .0013 & \cellcolor{grey!25} .8632 & \cellcolor{grey!25} .0014 & SN: Subject Number \\
\cline{1-11}
 abl. N & .9384 & .0005 & .994 & .0001 & .8058 & .0016 & .7743 & .0018 & .8594 & .0013 & ON: Object Number \\
 abl. D & \textbf{.5481} & .0013 & \textbf{.504} & .0011 & \cellcolor{cadetgrey!25} .5003 & \cellcolor{cadetgrey!25} .0022 & \cellcolor{cadetgrey!25} .4994 & \cellcolor{cadetgrey!25} .0024 & \cellcolor{cadetgrey!25} .5013 & \cellcolor{cadetgrey!25} .0025 & TE: Tense \\
 abl. D+N & \cellcolor{cadetgrey!25} .5001 & \cellcolor{cadetgrey!25} .0011 & \cellcolor{cadetgrey!25} .4999 & \cellcolor{cadetgrey!25} .0008 & \cellcolor{cadetgrey!25} .4987 & \cellcolor{cadetgrey!25} .0024 & \cellcolor{cadetgrey!25} .4994 & \cellcolor{cadetgrey!25} .002 & \cellcolor{cadetgrey!25} .4998 & \cellcolor{cadetgrey!25} .0021 & \textit{Syntax} \\
\cline{1-11}
Model & \multicolumn{2}{m{3.6em}|}{\textbf{CIN}} & \multicolumn{2}{m{3.8em}|}{\textbf{TD}} & \multicolumn{2}{m{3.4em}|}{\textbf{TC}} & \multicolumn{2}{m{3.8em}|}{\textbf{BS}} & \multicolumn{2}{m{4.5em}|}{\textbf{SOMO}} & CIN: Coordination \\
 & auc & \textpm CI & auc & \textpm CI & auc & \textpm CI & auc & \textpm CI & auc & \textpm CI & Inversion \\
\cline{1-11}
rand. pred. & \cellcolor{cadetgrey!25} .5004 & \cellcolor{cadetgrey!25} .0022 & \cellcolor{cadetgrey!25} .5005 & \cellcolor{cadetgrey!25} .0012 & \cellcolor{cadetgrey!25} .5005 & \cellcolor{cadetgrey!25} .0009 & \cellcolor{cadetgrey!25} .4998 & \cellcolor{cadetgrey!25} .0022 \cellcolor{cadetgrey!25} & .4999 & \cellcolor{cadetgrey!25} .0026 & TD: Parse Tree Depth \\
rand. vec. & \cellcolor{cadetgrey!25} .4993 & \cellcolor{cadetgrey!25} .0022 & \cellcolor{cadetgrey!25} .5002 & \cellcolor{cadetgrey!25} .0014 & \cellcolor{cadetgrey!25} .5004 & \cellcolor{cadetgrey!25} .0009 & \cellcolor{cadetgrey!25} .4989 & \cellcolor{cadetgrey!25} .0023 & \cellcolor{cadetgrey!25} .4991 & \cellcolor{cadetgrey!25} .0023 & TC: Top Constituents \\
\cline{1-11}
vanilla & \cellcolor{grey!25} .5493 & \cellcolor{grey!25} .0019 & \cellcolor{grey!25} .7799 & \cellcolor{grey!25} .0012 & \cellcolor{grey!25} .9512 & \cellcolor{grey!25} .0004 & \cellcolor{cadetgrey!25} .5017 & \cellcolor{cadetgrey!25} .0021 & \cellcolor{grey!25} .5291 & \cellcolor{grey!25} .0021 & \textit{Incongruity} \\
\cline{1-11}
abl. N  & .5437 & .002 & .7689 & .001 & .9438 & .0004 & \cellcolor{cadetgrey!25} .5034 & \cellcolor{cadetgrey!25} .0024 & .5235 & .002 & BS: Bigram Shift \\
abl. D & \cellcolor{cadetgrey!25} .5003 & \cellcolor{cadetgrey!25} .0023 & \textbf{.5137} & .0012 & \textbf{.5331} & .0013 & \cellcolor{cadetgrey!25} .499 & \cellcolor{cadetgrey!25} .0026 & \cellcolor{cadetgrey!25} .5005 & \cellcolor{cadetgrey!25} .0021 & SOMO: Semantic \\
abl. D+N & \cellcolor{cadetgrey!25} .5004 & \cellcolor{cadetgrey!25} .0021 & \cellcolor{cadetgrey!25} .501 & \cellcolor{cadetgrey!25} .0013 & \cellcolor{cadetgrey!25} .4996 & \cellcolor{cadetgrey!25} .0011 & \cellcolor{cadetgrey!25} .4996 & \cellcolor{cadetgrey!25} .0024 & \cellcolor{cadetgrey!25} .5007 & \cellcolor{cadetgrey!25} .0019 & Odd Man Out \\
\hline 
\end{tabular}}
\caption{Experimental results on GloVe models and baselines. Reporting average AUC-ROC scores and confidence intervals (CI) of the average of all training runs.
Cells shaded light grey belong to the same distribution as random baselines, dark grey cells share the vanilla baseline distribution, while scores significantly different from both the random and vanilla baselines are unshaded, while the most pertinent scores are marked in bold.}
\label{tab:results-glove}
\end{table*}

\subsection{Chosen Noise Models}\label{sub:noisemodels}

As described in \S\ref{s:method}, we remove information from the norm by sampling random norm values and scaling the vector dimensions to the new norm. However, considering that vectors have more than one calculable norm, the scaling can be done to match more than one norm value. We have examined the effects of scaling to both the L1 and L2 norms, as they are most widely used in NLP, and found that applying our norm ablation noise function to scale to either norm removes information from both norms (see Table \ref{tab:correlation}).\footnote{This contrasts with applying a normalisation function to the vector, where normalising to one of the norms removes information encoded in that norm, but retains, or even emphasises the information in the remaining norm, making normalisation an unsuitable ablation function (see \S\ref{sec:appendix_a} for details).} In order to streamline the results presentation, henceforth when discussing norm ablations we only report results pertaining to scaling to the L2 norm.

To ablate information encoded in the dimension container, we randomly sample dimension values and then scale them to match the original norm of the vector (see \S\ref{s:method}).\footnote{The random norm and dimension values are sampled uniformly from a range between the minimum and maximum norm/dimension values of the respective embeddings on all 10 datasets. BERT norm range: [7.1896,13.2854], BERT dimension range: [-5.427,1.9658]; GloVe norm range: [2.0041,8.0359], GloVe dimension range: [-2.5446,3.1976]} We expect this to fully remove all interpretable information encoded in the dimension values, making the norm the only information container available to the probe. Applying both noise functions together on the same vector should remove any information encoded in it.

Finally, we use the vanilla BERT and GloVe sentence embeddings in their respective evaluations as vanilla baselines against which the models with noise are compared. Here the probe has access to both information containers: dimensions and norm. However, it is also important to establish the vanilla baseline's performance against the random baselines: we need to confirm whether the information is in fact encoded somewhere in the embeddings.

\subsection{Results}\label{ss:results}

Detailed experimental evaluation results for GloVe and BERT on each of the 10 probing tasks are presented in Tables \ref{tab:results-glove} and \ref{tab:results-bert} respectively. Note that all cells shaded light grey belong to the same distribution as random baselines on a given task, as there is no statistically significant difference between the different scores\footnote{We highlight that the \textit{rand. vec.} baseline is equivalent to the scenario where both dimensions and norm are ablated (\textit{abl. D+N}). While the two scenarios are arguably the exact same condition, we include both of them in the results presentation to demonstrate a consistent application of our methodology, where we consider \textit{rand. vec.} to be a baseline, and the \textit{abl. D+N} a sense-check of our ablation functions.}; cells shaded dark grey belong to the same distribution as the vanilla baseline on a given task; and all cells that are not shaded contain a significantly different score than both the random and vanilla baselines, indicating that they belong to different distributions. The scores most pertinent to the result discussion are marked in bold.

\begin{table*}
\centering
\scalebox{.79}{
\centering
\begin{tabular}{|l|c|c|c|c|c|c|c|c|c|c|c|}
\hline
\multicolumn{11}{|c|}{\textbf{BERT}} & \textbf{Key} \\
\cline{1-11}
Model & \multicolumn{2}{m{3.6em}|}{\textbf{SL}} & \multicolumn{2}{m{4.1em}|}{\textbf{WC}} & \multicolumn{2}{m{3.8em}|}{\textbf{SN}} & \multicolumn{2}{m{3.8em}|}{\textbf{ON}} & \multicolumn{2}{m{3.8em}|}{\textbf{TE}} & \textit{Surface Info.} \\
 & auc & \textpm CI & auc & \textpm CI & auc & \textpm CI & auc & \textpm CI & auc & \textpm CI & SL: Sentence Length \\
\cline{1-11}
rand. pred. & \cellcolor{cadetgrey!25} .5002 & \cellcolor{cadetgrey!25} .0006 & \cellcolor{cadetgrey!25} .4996 & \cellcolor{cadetgrey!25} .0012 & \cellcolor{cadetgrey!25} .4995 & \cellcolor{cadetgrey!25} .0021 & \cellcolor{cadetgrey!25} .4988 & \cellcolor{cadetgrey!25} .0022 & \cellcolor{cadetgrey!25} .5007 & \cellcolor{cadetgrey!25} .0021 & WC: Word Content \\
rand. vec. & \cellcolor{cadetgrey!25} .5003 & \cellcolor{cadetgrey!25} .0004 & \cellcolor{cadetgrey!25} .4997 & \cellcolor{cadetgrey!25} .0009 & \cellcolor{cadetgrey!25} .5006 & \cellcolor{cadetgrey!25} .002 & \cellcolor{cadetgrey!25} .4996 & \cellcolor{cadetgrey!25} .0024 & \cellcolor{cadetgrey!25} .4993 & \cellcolor{cadetgrey!25} .0021 & \textit{Morphology} \\
\cline{1-11}
vanilla & \cellcolor{grey!25} .9733 & \cellcolor{grey!25} .0011 & \cellcolor{grey!25} .982 & \cellcolor{grey!25} .0003 & \cellcolor{grey!25} .9074 & \cellcolor{grey!25} .0008 & \cellcolor{grey!25} .8674 & \cellcolor{grey!25} .0019 & \cellcolor{grey!25} .9135 & \cellcolor{grey!25} .0008 & SN: Subject Number \\
\cline{1-11}
abl. N & \cellcolor{grey!25} .973 & \cellcolor{grey!25} .0008 & .9783 & .0003 & \cellcolor{grey!25} .9078 & \cellcolor{grey!25} .0008 & \cellcolor{grey!25} .8658 & \cellcolor{grey!25} .0017 & \cellcolor{grey!25} .9118 & \cellcolor{grey!25} .0012 & ON: Object Number \\
abl. D & \cellcolor{cadetgrey!25} .5047 & \cellcolor{cadetgrey!25} .0008 & \cellcolor{cadetgrey!25} .5013 & \cellcolor{cadetgrey!25} .0011 & \cellcolor{cadetgrey!25} .4992 & \cellcolor{cadetgrey!25} .0021 & \cellcolor{cadetgrey!25} .5004 & \cellcolor{cadetgrey!25} .0023 & \cellcolor{cadetgrey!25} .5007 & \cellcolor{cadetgrey!25} .0019 & TE: Tense \\
abl. D+N & \cellcolor{cadetgrey!25} .4997 & \cellcolor{cadetgrey!25} .0008 & \cellcolor{cadetgrey!25} .5 & \cellcolor{cadetgrey!25} .0013 & \cellcolor{cadetgrey!25} .5006 & \cellcolor{cadetgrey!25} .0024 & \cellcolor{cadetgrey!25} .4994 & \cellcolor{cadetgrey!25} .0024 & \cellcolor{cadetgrey!25} .4983 & \cellcolor{cadetgrey!25} .0021 & \textit{Syntax} \\
\cline{1-11}
\cline{1-11}
Model & \multicolumn{2}{m{3.6em}|}{\textbf{CIN}} & \multicolumn{2}{m{3.8em}|}{\textbf{TD}} & \multicolumn{2}{m{3.4em}|}{\textbf{TC}} & \multicolumn{2}{m{3.8em}|}{\textbf{BS}} & \multicolumn{2}{m{4.5em}|}{\textbf{SOMO}} & CIN: Coordination \\
 & auc & \textpm CI & auc & \textpm CI & auc & \textpm CI & auc & \textpm CI & auc & \textpm CI & Inversion \\
\cline{1-11}
rand. pred. & \cellcolor{cadetgrey!25} .5007 & \cellcolor{cadetgrey!25} .0022 & \cellcolor{cadetgrey!25} .4999 & \cellcolor{cadetgrey!25} .0012 & \cellcolor{cadetgrey!25} .5001 & \cellcolor{cadetgrey!25} .0013 & \cellcolor{cadetgrey!25} .5011 & \cellcolor{cadetgrey!25} .0020 & \cellcolor{cadetgrey!25} .499 & \cellcolor{cadetgrey!25} .0018 & TD: Parse Tree Depth \\
rand. vec. & \cellcolor{cadetgrey!25} .5014 & \cellcolor{cadetgrey!25} .0019 & \cellcolor{cadetgrey!25} .4999 & \cellcolor{cadetgrey!25} .0012 & \cellcolor{cadetgrey!25} .5001 & \cellcolor{cadetgrey!25} .0013 & \cellcolor{cadetgrey!25} .5005 & \cellcolor{cadetgrey!25} .0024 & \cellcolor{cadetgrey!25} .5001 & \cellcolor{cadetgrey!25} .0021 & TC: Top Constituents \\
\cline{1-11}
vanilla & \cellcolor{grey!25} .7472 & \cellcolor{grey!25} .0016 & \cellcolor{grey!25} .7751 & \cellcolor{grey!25} .0016 & \cellcolor{grey!25} .9562 & \cellcolor{grey!25} .0002 & \cellcolor{grey!25} .9382 & \cellcolor{grey!25} .0006 & \cellcolor{grey!25} .6401 & \cellcolor{grey!25} .0013 & \textit{Incongruity} \\
\cline{1-11}
abl. N & \cellcolor{grey!25} .7492 & \cellcolor{grey!25} .0018 & .7709 & .0016 & .9547 & .0004 & \cellcolor{grey!25} .9371 & \cellcolor{grey!25} .001 & \cellcolor{grey!25} .6396 & \cellcolor{grey!25} .0017 & BS: Bigram Shift \\
abl. D  & \cellcolor{cadetgrey!25} .5049 & \cellcolor{cadetgrey!25} .0021 & \cellcolor{cadetgrey!25} .5004 & \cellcolor{cadetgrey!25} .0013 & \textbf{.5093} & .0019 & \textbf{.556} & .0025 & \textbf{.5272} & .002 & SOMO: Semantic \\
abl. D+N & \cellcolor{cadetgrey!25} .5015 & \cellcolor{cadetgrey!25} .0035 & \cellcolor{cadetgrey!25} .5 & \cellcolor{cadetgrey!25} .0012 & \cellcolor{cadetgrey!25} .5001 & \cellcolor{cadetgrey!25} .001 & \cellcolor{cadetgrey!25} .4972 & \cellcolor{cadetgrey!25} .0035 & \cellcolor{cadetgrey!25} .4997 & \cellcolor{cadetgrey!25} .002 & Odd Man Out \\
\hline
\end{tabular}}
\caption{Experimental results on BERT models and baselines. Reporting average AUC-ROC scores and confidence intervals (CI) of the average of all training runs.
Cells shaded light grey belong to the same distribution as random baselines, dark grey cells share the vanilla baseline distribution, while scores significantly different from both the random and vanilla baselines are unshaded, while the most pertinent scores are marked in bold.}
\label{tab:results-bert}
\end{table*}

\textbf{GloVe results:}
The vanilla GloVe vectors outperform the random baselines on all tasks except BS. This is not surprising, as BS is essentially a local-context task, and GloVe does not encode context in such a localised manner. In all other tasks, at least some task-relevant information is encoded in the embeddings. Having established the vanilla results as a baseline for the ablations, we examine which information container encodes the relevant information: dimension or norm. 

Generally, the results show that the answers are task-dependent. In the SN, ON, TE, CIN and SOMO tasks, there is a substantial drop in the probe's performance after ablating the dimension container and it is immediately comparable to random baselines. Furthermore, performance does not significantly change after also ablating the norm, indicating that for these tasks no pertinent information is stored in the norm, and that all the information the probe uses is stored in the dimensions.

However, the results for the SL, WC, TD and TC probes tell a different story. Once the dimension container is ablated from these vectors, although the performance drops markedly compared to vanilla, it does not quite reach the random baseline performance as observed in the above tasks.\footnote{This is true even in the case of WC, where the difference is really quite small, yet still statistically significant. Note that the WC task is a particularly unusual classification task, as there are 1000 possible classes to predict, which could explain the statistical significance of such a small difference.} These results indicate that for these tasks the relevant information is not contained \textit{only} in the dimension container. Furthermore, when the dimension and norm ablation functions are applied together, this induces a further performance drop, and the resulting performance scores become comparable to the random baselines. This indicates that the vectors with ablated dimension information still contain residual information relevant to the task, which is removed when also ablating the norm, pointing to the fact that the norm contains some of the relevant information \textit{regardless of what is encoded in the vector dimensions}.

We should note here that, while it is true that in all tasks ablating the norm alone causes a statistically significant drop in performance, this finding on its own should not be taken as an indicator that the norm encodes task-relevant information. Given how consistently small the drop is across all tasks ($<$0.1), this is more likely an artefact of an interaction between the noising function and the GloVe vectors. The more reliable indicator of where the information is encoded is the experiment on dimension ablations compared to ablating both dimension and norm: if for a particular task performance remains above random after ablating dimensions, but drops to random when ablating both dimensions and norms, this shows that the norm is encoding at least part of the relevant information.

\textbf{BERT results:}
The vanilla BERT vectors outperform random baselines across all tasks, including the BS task. When ablating the dimensions on most tasks, the probe's performance drops dramatically and is comparable to random baselines. It does not change after also ablating the norm, indicating that no pertinent information is stored in BERT's norm container for these tasks. However, the BS and SOMO tasks show that some of the task information is stored in BERT's norm, as the performance drop when ablating dimensions is not comparable to random baselines, and only reaches that once the norm is also ablated. The same is true for the syntactic TC task, which is also the only BERT result that shows a similar trend as GloVe, though it seems that BERT stores far less TC information in the norm than GloVe does.

Ultimately, our experimental results allow us to make a number of general inferences: (a) the norm is indeed a separate information container, (b) on most tasks the vast majority of the relevant information is encoded in the dimension values, but can be supplemented with information from the norm, (c) though the information contained in the norm is not always very impactful, it is not negligible, (d) different encoders use the norm to carry different types of information, (e) specifically BERT stores information pertinent to the BS, SOMO and TC tasks in the norm, (f) while GloVe uses it to store SL, WC, TC and TD information.

\subsection{Norm Correlation Analysis}\label{ss:norm}

While we have demonstrated that information can be encoded in the norm, we wish to also understand the relationship between the norms and the probed information. We explore this with a Pearson correlation analysis: we test the correlation between each vector norm and the sentence labels on each probing task dataset.\footnote{The Pearson test only works on continuous variables, but it is still possible to calculate with categorical variables if they are binary, by simply converting the categories to 0 and 1.} The correlation results are presented in Table \ref{tab:correlation}, and largely support our result interpretations from \S\ref{ss:results},\footnote{In cases such as WC and TC where there are more than two categorical variables we can perform a Kruskal-Wallis test to determine a statistically significant difference between the categories. This does not quantify the difference in the same way as a Pearson test, and does not allow us to determine whether the correlation is positive or not, nor how strong it is. Instead we can only say that we performed the test and found the results to be significant, indicating some correlation.} including that applying our noise function to ablate the norm fully removes the information from the norms: the correlation between either norm and the class labels drops to $\approx$0,\footnote{Except in GloVe-SL-L1 where the coefficient 'only' drops from strongly correlated to weakly correlated.} indicating that information encoded by the norm and any distinguishing properties it may have had have been removed.

\begin{table}[tb]
\centering
\scalebox{0.80}{
\centering
\begin{tabular}{|l|l|c|c|c|c|}
\hline
Task & Vectors & \multicolumn{2}{m{5.5em}|}{\textbf{GloVe}} & \multicolumn{2}{m{5.5em}|}{\textbf{BERT}} \\
 & & L1 & L2 & L1 & L2 \\
\hline
\hline
 & Vanilla & -0.7278 & -0.3758 & -0.1564 & -0.1039 \\
SL & Abl. norm & -0.1893 & -0.0025 & -0.0417 & -0.0013 \\
\hline
\hline
 & Vanilla & 0.0360 & 0.0268 & 0.0071 & 0.0146 \\
SN & Abl. norm & 0.0036 & -0.0033 & -0.0035 & -0.0021 \\
\hline
\hline
 & Vanilla & 0.0013 & 0.0008 & -0.0736 & -0.0583 \\
ON & Abl. norm & 0.0009 & 0.0013 & -0.0181 & -0.0010 \\
\hline
\hline
 & Vanilla & 0.1152 & 0.0571 & 0.0542 & 0.0413 \\
TE & Abl. norm & 0.0277 & -0.0031 & 0.0097 & -0.0030 \\
\hline
\hline
 & Vanilla & -0.0817 & 0.1908 & -0.0415 & -0.0251 \\
TD & Abl. norm & -0.0665 & 0.0016 & -0.0163 & -0.0045 \\
\hline
\hline
& Vanilla & -0.0019 & -0.0094 & -0.0755 & -0.0638 \\
CIN & Abl. norm & 0.0029 & 0.0018 & -0.0152 & -0.0015 \\
\hline
\hline
 & Vanilla & 0.0040 & 0.0002 & -0.3866 & -0.3238 \\
BS & Abl. norm & 0.0022 & 0.0006 & -0.0978 & -0.0005 \\
\hline
\hline
SO & Vanilla & -0.0464 & -0.0222 & -0.2414 & -0.2305 \\
MO & Abl. norm & -0.0105 & 0.0000 & -0.0420 & 0.0021 \\

\hline
\end{tabular}}
\caption{Pearson correlation coefficients between the class labels and vector norms for vanilla vectors and vectors with ablated norms.} 
\label{tab:correlation}
\end{table}

The data shows that most task labels do not exhibit a correlation with the vanilla GloVe norm. There is a moderate positive correlation between TD and the L2 norm, but not the L1 norm, and a weak positive correlation between TE and the L1 norm, but not the L2 norm. There is a high correlation between the SL labels and both norms, showing that GloVe uses the norm to encode sentence length, as reflected in our experiments in \S\ref{s:experiments}.

When it comes to vanilla BERT, most task labels do not exhibit a correlation with the norms. However, both norms have a weak negative correlation with SL, and a moderate negative correlation with BS and SOMO. The latter two are most highly correlated with BERT's norm, which also aligns with our experimental findings in \S\ref{s:experiments}.

\section{Discussion}\label{s:discussion}

The correlation coefficients in Table \ref{tab:correlation} can be interpreted in terms of how these linguistic phenomena are encoded in vector space. A negative correlation coefficient means that larger norms indicate a negative class, while a positive coefficient means that larger norms indicate a positive class. For example, the negative correlation in SL-GloVe and SL-BERT indicates that longer sentences are positioned closer to the origin. The same interpretation holds for BERT embeddings on the BS and SOMO tasks; e.g. in SOMO a sentence containing an out of context word is positioned closer to the origin.

It is interesting that BERT's norm stores information on the BS and SOMO tasks specifically. Their common thread is a violation of the local context of the affected words: though the overall context and structure of the sentence is unaffected, there is a small, localised disruption in co-occurrences. Hence, these tasks capture contextual incongruity. Given that we know that BERT is a contextual encoder, and that its self-attention uses the vector norm to control the levels of contribution from less informative words \citep{kobayashi-etal-2020-attention}, we suspect that this gives it the capabilities to accurately model these short-distance dependencies and word co-occurence probabilities, concepts which strongly correspond to local contextual incongruity. BERT is evidently capable of encoding this signal well, and seems to be using its norm to supplement the encoding of the phenomenon in such a way that it positions sentences exhibiting local contextual incongruity closer to the origin, relative to sentences that do not contain it. 
Furthermore, BERT's ability to model incongruity via the norm could essentially be frequency-based, similar to how some word embeddings encode word frequency in the norm.
In contrast, GloVe is a static encoder and exhibits no indication that it stores this information in the norm, or indeed any ability to accurately model this phenomenon at all, but uses the norm to store surface-level and syntactic information.

We emphasise the importance of the norm as it expands our understanding of the way information is encoded in vector space, but it could also have important implications for downstream tasks involving operations on vectors: e.g. the calculation of a cosine similarity measure normalises the vectors being compared. This nullifies the information in the norm, reducing the comparison to one of directions (i.e. dimensions), and any linguistic information encoded in the norm will be lost and unaccounted for when making the comparison.

\section{Related Work}
\label{s:rw}

Probing has been proposed seemingly independently by different groups of NLP researchers \cite{ettinger-etal-2016-probing,shi-etal-2016-string,veldhoen2016diagnostic,adi2017fine} and has gained significant momentum in the community, helping to explore different aspects of text encodings (e.g. \citet{hupkes2018visualisation,giulianelli2018under,krasnowska-kieras-wroblewska-2019-empirical,tenney-etal-2019-bert,lin2019open,sahin-etal-2020-linspector,liu-etal-2021-probing-across,arps-etal-2021-probing}). Probes trained on various representations successfully predict surface properties of sentences \citep{adi2017fine,Conneau:2018}, POS and morphological information \citep{belinkov-etal-2017-neural,liu-etal-2019-linguistic}, as well as syntactic \citep{zhang-bowman-2018-language,peters2018dissecting,tenney2019you}, semantic \cite{belinkov-etal-2017-evaluating,ahmad2018multitask,conia-navigli-2022-probing}, and even number \citep{wallace-etal-2019-nlp}, emotions \cite{qian-etal-2016-investigating}, idiomaticity \citep{salton-etal-2016-idiom,nedumpozhimana-kelleher-2021-finding,garcia-etal-2021-probing,nedumpozhimana-etal-2022-shapley} and world knowledge information \citep{ettinger2020bert}, among others \cite{belinkov-glass-2019-analysis,rogers2020primer,koto-etal-2021-discourse,ousidhoum-etal-2021-probing,aghazadeh-etal-2022-metaphors}.

Furthermore, some dichotomies have emerged in the literature, due to nuanced differences in the presuppositions behind probing approaches. \citet{ravichander2020systematicity} distinguish varying points of view on embeddings, highlighting a difference between \textit{instrumentative} and \textit{agentive} probing. \citet{vig2020causal} view probing as a method of analysis and distinguish two types of methods: \textit{structural} and \textit{behavioural}. Additionally, \citet{pimentel-etal-2020-information} and \citet{voita-titov-2020-information} take an information-theoretic perspective on embeddings, highlighting the tension between probing identifying the mere \textit{presence} of information, versus its \textit{extractability}. We position our work as being \textbf{instrumentative}, i.e. we view embeddings as tools that extract and store knowledge from text; we consider our probing method to be \textbf{structural}, i.e. it provides insight into how information is encoded within the representation and the vector space; and the goal of our work is to identify the \textbf{presence} of information in embedding components. It is important to clearly signpost this position in order to avoid confusion and emphasise that our chosen approach is sufficient to address our research questions.

Meanwhile, recent work calls for greater rigor in evaluation approaches in NLP \cite{mccoy-etal-2020-berts,sadeqi-azer-etal-2020-claims,card-etal-2020-little}, advocating for more widespread use of statistical tests on common benchmarks. Probing has attracted similar criticism: \citet{hewitt-liang-2019-designing} have shown that under certain conditions, above-random probing accuracy can be achieved even when probing for linguistically-meaningless noise. 
Recent work addresses some of these problems by constructing counterfactual representations in order to compare the performance of the probe with and without the pertinent information \cite{feder2020causalm}. Similarly, \citet{elazar2020bert} remove the relevant information from the representation, allowing a comparison of probe performance with and without the removed information; not unlike the intrinsic probe of \citet{torroba-hennigen-etal-2020-intrinsic} who focus on isolating the dimensions that encode relevant information. In essence, these recent efforts address the issue of relativising probe interpretations by removing information from the encoding; in that sense, our work finds its place alongside them. However, our method is not meant to remove specific information, but is more exploratory in nature, with a focus on understanding where within an embedding certain information is encoded. Our use of confidence intervals gives us a way to claim statistically significant differences in our evaluations, offering a more principled basis for result interpretation.

Our work also contributes to the relatively scarce study of the role of the norm: \citet{adi2017fine} explain its correlation with SL information due to the central limit theorem (which we see does not apply to BERT as its vector values are not centred around zero). \citet{hewitt-manning-2019-structural} show that the squared L2 norm of BERT and ELMo corresponds to the depth of the word in a parse tree (a finding we could not confirm as they probe embeddings at the word level, unlike our work). In contrast, work on the role of dimensions as carriers of specific types of information is plentiful (e.g. \citet{karpathy-etal-2015-visualising,qian-etal-2016-investigating,bau2019identifying,dalvi-etal-2019-everything,lakretz-etal-2019-emergence}). Work complementary to ours  \cite{torroba-hennigen-etal-2020-intrinsic} which focuses on the dimension container also highlights the need for an intrinsic probe of embedding models, and shows that most linguistic properties are reliably encoded by only a handful of dimensions, a finding consistent with \citet{durrani-etal-2020-analyzing} and \citet{durrani2022linguistic}.

\section{Conclusion}

We have developed a method of enquiry that provides geometric insights into embeddings and show experimental evidence that both BERT and GloVe embeddings use two separate information containers to store different types of linguistic information. Our findings show that BERT primarily uses the norm to store contextual incongruity information and positions incongruous sentences closer to the origin. Meanwhile, GloVe stores much more syntactic information in its norm than BERT, but does not store contextual information at all, and mainly stores surface-level information in the norm.

\textit{Probing with noise} can shift perspectives and broaden our understanding of embeddings, demonstrated by our experiments which provide novel insights into contextual and static encoders. However, they are by no means exhaustive: deeper and further applications of the method, such as exploring a host of other representations, different pooling strategies or tracking behavior across embedding layers, exploring word-level tasks or folding in additional datasets, are all fruitful avenues for future work. Fortunately, the method is robust enough to be applied to any encoder and any dataset, whether it is at the word or sentence level, which allows for systematic further study.

\section{Limitations}

While our insights into how linguistic information can be encoded in embeddings are valuable on their own merit, our experiments mainly serve the purpose of validating the \textit{probing with noise} method, in demonstrating that it can produce relevant insights on different types of embeddings. Hence we did not have scope to more thoroughly pursue many of the topics touched upon in the paper. 

One example is our choice in generating sentence embeddings needed to probe for sentence-level information. The encoders we have used generate word-level embeddings, so we average the word embeddings in each sentence, as this is one of the most popular ways to generate sentence representations. However, there are other known approaches available to choose from, such as max pooling and min pooling, or, when it comes to BERT, using the CLS token.\footnote{Presumably, we may have observed a crisper effect in BERT encoding incongruity using min or max pooling, given that the BS task mainly affects only a few vectors in a sentence.} Indeed, rather than a pooling strategy, using direct sentence-level representations such as doc2vec \cite{le2014distributed} or SentenceBERT \cite{reimers-gurevych-2019-sentence} might also be prudent, as well as applying the method to word-level representations, for which this paper did not allow scope. 

Similarly, we have consistently used only one probing classifier, an MLP with default parameters, and we cannot say whether parameter tuning or different probes would yield different results. These choices were made consciously, in order to avoid adding more variables to our line of enquiry and increasing the complexity of our experiments, yet it is still a limitation in the sense that we do not know whether the findings generalise to other probes.

It is also worth noting that the correlation study in \S\ref{s:experiments} comes with the limitation of only describing linear relationships, whereas it is possible that connections between variables can be non-linear. We argue that this demonstrates the value of our method, which allows for a non-linear probe to test for non-linear relationships. While even this limited correlation test can provide interesting insights, much more can be done to study both the norm and the dimension container---we have just barely scratched the surface. Indeed, we have considered only the most fundamental geometric properties of vectors, yet vectors have other (distributed) properties that could potentially be considered distinct information containers in their own right, such as the vector's minimum and maximum value, their ratio, the entropy in the vector etc. Thankfully the principles underpinning our method can be expanded to include other types of noise that help discriminate other possible geometric properties of embeddings as information containers.

These points speak to the more general limitations of our research: like any empirical work, we measure behaviours on a number of data points and draw conclusions from these measurements. Thus there is a risk that our findings hold only for the datasets on which we measured or the models which were used to measure, be it encoders, probes or probing tasks, and it is possible that our findings might not generalise to other settings. While this issue is more epistemological than it is specific to our work, we must keep it in mind. Now, having demonstrated that a signal is detectable in our particular setting, a more comprehensive host of studies is needed to draw more general conclusions. 

Another source of uncertainty stems from our use of off-the-shelf GloVe and BERT embeddings: they have been trained on completely different datasets of dramatically varying sizes and content. To truly test the interaction of their architectures with our method, the training data used to train their word embeddings should be identical between both encoders, however implementing this was not feasible in practice. Granted, using off-the-shelf varieties does provide insight into the functioning of well-known and commonly used embeddings, but it consequently limits the comparability of their results as we cannot confidently distinguish whether differences in probe performance are due to differences in encoder architecture or training data.

While we acknowledge a number of the work's limitations, we stress that all our choices have been made in a sound, informed and methodologically consistent manner. Here we simply highlight just how many choices have been made along the way, and how quickly the number of alternative paths grows the further back up the decision tree we look. While we believe that the work is fundamentally sound, each choice could have made for a drastically different suite of experiments and could potentially have yielded different results. In fact, we find this to be a very exciting motivator for future work, as this long list of ``missed opportunities'' only goes to show how young and rich this research area still is and how many more avenues there are to explore.

\section*{Acknowledgements}

This research was conducted with the financial support of Science Foundation Ireland under Grant Agreements No. 13/RC/2106 and 13/RC/2106\_P2 at the ADAPT SFI Research Centre at Technological University Dublin. ADAPT, the SFI Research Centre for AI-Driven Digital Content Technology, is funded by Science Foundation Ireland through the SFI Research Centres Programme, and is co-funded under the European Regional Development Fund.

\bibliography{anthology,bbnlp2022-f}
\bibliographystyle{acl_natbib}

\appendix

\section{Appendix A}
\label{sec:appendix_a}

\subsection{Analysis of L1 and L2 Normalised Embeddings}

\begin{table}[tb]
\centering
\scalebox{0.80}{
\centering
\begin{tabular}{|l|l|c|c|c|c|}
\hline
Task & Vectors & \multicolumn{2}{m{5.5em}|}{\textbf{GloVe}} & \multicolumn{2}{m{5.5em}|}{\textbf{BERT}} \\
 & & L1 & L2 & L1 & L2 \\
\hline
\hline
 & Vanilla & -0.7278 & -0.3758 & -0.1564 & -0.1039 \\
SL & L1 normal. & -0.0013	 & 0.7161 & 0.0032	& 0.2195 \\ 
 & L2 normal. & -0.7027 & 0.0001 &-0.2223 & 0.0001 \\ 
 & Abl. norm & -0.1893 & -0.0025 & -0.0417 & -0.0013 \\
\hline
\hline
 & Vanilla & 0.0360 & 0.0268 & 0.0071 & 0.0146 \\
SN & L1 normal. & 0.0028 & -0.0228  & -0.0010 & 0.0087 \\ 
 & L2 normal. & 0.0255 & -0.0019 & -0.0086 & -0.0003 \\ 
 & Abl. norm & 0.0036 & -0.0033 & -0.0035 & -0.0021 \\
\hline
\hline
 & Vanilla & 0.0013 & 0.0008 & -0.0736 & -0.0583 \\
ON & L1 normal. & -0.0016	 & 0.0048 & -0.0015 & 0.0892 \\ 
 & L2 normal. & -0.0004 & -0.0015 & -0.0901 & 0.0037 \\ 
 & Abl. norm & 0.0009 & 0.0013 & -0.0181 & -0.0010 \\
\hline
\hline
 & Vanilla & -0.1152 & -0.0571 & -0.0542 & -0.0413 \\
TE & L1 normal. & -0.0020 & 0.1040 & -0.0023 & 0.0659 \\ 
 & L2 normal. & -0.1071 & -0.0006 & -0.0691 & -0.0018 \\ 
 & Abl. norm & -0.0317 & -0.0007 & -0.0116 & 0.0010 \\
\hline
\hline
 & Vanilla & -0.0817 & 0.1908 & -0.0415 & -0.0251 \\
TD & L1 normal. & 0.0005 & 0.3133 & 0.0021 & 0.0645 \\ 
 & L2 normal. & -0.3159 & -0.0026 & -0.0652 & 0.0000 \\ 
 & Abl. norm & -0.0665 & 0.0016 & -0.0163 & -0.0045 \\
\hline
\hline
 & Vanilla & -0.0019 & -0.0094 & -0.0755 & -0.0638 \\
CIN & L1 normal. & 0.0000 & -0.0062 & -0.0047 & 0.0846 \\ 
 & L2 normal. & 0.0065 & 0.0064 & -0.0850 & 0.0034 \\ 
 & Abl. norm & 0.0029 & 0.0018 & -0.0152 & -0.0015 \\
\hline
\hline
 & Vanilla & 0.0040 & 0.0002 & -0.3866 & -0.3238 \\
BS & L1 normal. & -0.0015	 & -0.0048  & 0.0004 & 0.4333 \\ 
 & L2 normal. & 0.0056 & -0.0019 & -0.4357 & 0.0024 \\ 
 & Abl. norm & 0.0022 & 0.0006 & -0.0978 & -0.0005 \\
\hline
\hline
 & Vanilla & -0.0464 & -0.0222 & -0.2414 & -0.2305 \\
SO & L1 normal. & 0.0031 & 0.0401 & 0.0035 & 0.2213 \\ 
MO & L2 normal. & -0.0392	 & -0.0014 & -0.2219 & 0.0023 \\ 
 & Abl. norm & -0.0105 & 0.0000 & -0.0420 & 0.0021 \\
\hline
\end{tabular}}
\caption{Pearson correlation coefficients between the class labels and vector norms for vanilla vectors, L1 and L2 normalised vectors, as well as vectors with ablated L2 norm containers.}
\label{tab:correlation_appendix}
\end{table}

Table \ref{tab:correlation_appendix} presents an extended Pearson correlation analysis that includes correlations between class labels and the norms of L1- and L2-normalised vectors, in addition to vanilla vectors and vectors with ablated norm information using our noising function as described in \S\ref{s:method}. 

As supported by \citet[page 117]{goldberg2017neural}, the results show that normalising the vectors removes information encoded in the norm. This also comes with a caveat: normalisation only removes information from the same order norm as the normalisation algorithm. We can observe this in the table: applying an L1 normalisation algorithm to the vectors seems to completely remove any information encoded in the L1 norm, as the correlation drops to $\approx0$. The same happens to the correlation with the L2 norm when applying L2 normalisation. However, surprisingly, it seems that a given normalisation algorithm impacts the other norm as well. For example, in the BS task L2 normalisation nullifies the L2 norm's correlation with the class labels, but in turn strengthens that correlation for the L1 norm, which intensifies from -0.39 to -0.44. On the other hand, L1 normalisation causes the same strengthening of correlation in the L2 norm, but also changes the sign---the L2 norm's correlation with BS class labels increases from -0.32 to 0.43.

This shows that on certain tasks, not only is the other norm unaffected by a normalisation procedure, but its correlation with the task labels increases. We observe this to varying degrees in SL, ON, TE and BS. Furthermore, while the correlation weakens in SOMO, it still exhibits the latter behaviour---the sign changes when the vectors are L1 normalised, but not when they are L2 normalised. This is prevalent across all datasets, even in cases where the correlation between norm and class labels is $\approx$0.

This analysis supports our decision from \S\ref{s:method} to use a different noising function to remove information from the norm container, as only the vectors with fully ablated norms have an $\approx$0 correlation with both the L1 and L2 norms.

\end{document}